\documentclass[10pt,twocolumn,letterpaper]{article}

\usepackage[pagenumbers]{cvpr}              % To produce the CAMERA-READY version
\usepackage{graphicx}
\usepackage{booktabs}
\usepackage{adjustbox}
\usepackage{multirow}

\newcommand{\blfootnote}[1]{%
    \begingroup
    \renewcommand\thefootnote{}\footnote{#1}%
    \addtocounter{footnote}{-1}%
    \endgroup
}
\usepackage[dvipsnames]{xcolor}

\definecolor{cvprblue}{rgb}{0.21,0.49,0.74}
\usepackage[pagebackref,breaklinks,colorlinks,citecolor=cvprblue]{hyperref}

\def\paperID{11} % *** Enter the Paper ID here
\def\confName{CVPR}
\def\confYear{2024}

\title{ViSpeR: Multilingual Audio-Visual Speech Recognition}

\author{Sanath Narayan$^{1}$\thanks{Joint first author; order determined by flipping a coin.}\hspace{0.5em}
 \quad Yasser Abdelaziz Dahou Djilali$^{1*}$ \hspace{0.5em}
 \quad Ankit Singh$^{1}$\\
Eustache Le Bihan$^{2}$\thanks{Work done during an internship at TII.} \hspace{1.5em} \quad Hakim Hacid$^{1}$ \\
\\
$^{1}$Technology Innovation Institute, UAE \hspace{2.5em}
$^{2}$ENS Paris-Saclay, France \\
}

\begin{document}

\maketitle
\begin{abstract}
This work presents an extensive and detailed study on Audio-Visual Speech Recognition (AVSR) for five widely spoken languages: Chinese, Spanish, English, Arabic, and French. We have collected large-scale datasets for each language except for English, and have engaged in the training of supervised learning models. Our model, ViSpeR, is trained in a multi-lingual setting, resulting in competitive performance on newly established benchmarks for each language. The datasets and models are released to the community with an aim to serve as a foundation for triggering and feeding further research work and exploration on Audio-Visual Speech Recognition, an increasingly important area of research. Code available at \href{https://github.com/YasserdahouML/visper}{https://github.com/YasserdahouML/visper}.

\blfootnote{Please send correspondence to: visper@tii.ae}
\end{abstract}    
\section{Introduction}
 
Visual Speech Recognition (VSR), also known as sentence-level VSR~\cite{Sheng2022DeepSurvey}, poses significant challenges in training deep learning models due to the ambiguous nature of the input data, and the lack of large scale datasets compared to, e.g., the ones in Audio Speech Recognition (ASR)~\cite{chen2021gigaspeech, ardila2019common}. Indeed, acquiring VSR data involves recording and annotating both the audio and video streams simultaneously, which is a more complex and resource-intensive process compared to acquiring only audio data for ASR. Additionally, factors like lighting conditions, camera angles, and speaker variations can introduce noise and variability in the visual data, making it harder to capture high-quality and consistent VSR data at scale. Furthermore, before extracting the visual features from a video stream, it is necessary to first detect the active speaker in the video and then locate and track the mouth region of the person of interest, which adds further complexity to the data acquisition process.

To this end, not only are the current VSR datasets (e.g., LRS3~\cite{afouras2018lrs3} and VoxCeleb~\cite{nagrani2020voxceleb}) smaller in size compared to ASR datasets, they are mostly focused on the English language. This limits the applicability of VSR models to other languages and accents. Indeed, existing non-English VSR datasets~\cite{zadeh2020cmu, ivanko2022rusavic} are significantly shorter in duration, and often recorded in a controlled environment. The recent work of MuAVIC~\cite{anwar2023muavic} introduced a multilingual audio-visual corpus, providing $1,200$ hours of audio-visual speech in nine languages extracted from TED talks. Authors of \cite{yeo2024visual} leveraged the existing large-scale, unlabeled multilingual audio-visual speech datasets, such as VoxCeleb2~\cite{nagrani2020voxceleb} (2442 hours) and AV-Speech~\cite{ephrat2018looking} (around $4,700$ hours), and used Whisper~\cite{radford2023robust} to transcribe the segments. This led to the creation of a multi-lingual dataset for four languages, namely: French, Italian, Spanish and Portuguese.

Given the scarcity of publicly available VSR data for non-English languages, the first natural step is to collect such data for the most four spoken languages at scale. To this end, we develop a data pipeline for efficiently collecting and processing videos from the wild. In this work, we collect and process data for Arabic (\textit{ar}), Spanish (\textit{es}), French (\textit{fr}), and Chinese (\textit{zh}).

\noindent\textbf{Contributions:} Based on the above discussion, the main contributions of this work are as follows:
\begin{itemize}
    \item \textit{Data pipeline and dataset}: We develop an efficient data collection pipeline for VSR to obtain 787h, 1200h, 794h, and 872h of data for Chinese, Arabic, Spanish, and French languages, respectively. 
    \item \textit{Benchmarks}: We also carefully create benchmarks for each language so they can be used to further measure progress in this field.  
    \item \textit{ViSpeR}: We engage in the training of supervised VSR and AVSR models, and establish them as baselines on the introduced benchmarks.
    
\end{itemize}

\section{Building the ViSpeR dataset}

Exploiting publicly available online content as a data source for creating audio/visual speech recognition datasets has become a popular approach ~\cite{makino2019recurrent, afouras2018lrs3, nagrani2020voxceleb, ephrat2018looking}. Clearly however, producing a VSR dataset of satisfactory quality necessitates meticulous processing of raw videos to create pairs of visual sequences, specifically visual lip movements matched with text labels. Due to the computational complexity, it is important to carefully filter and extract relevant content from the extensive online repository before proceeding. Therefore, the procedure should involve two primary stages: 1) identifying pertinent videos likely to contain VSR content, and 2) processing the selected videos to form the required data pairs.

In terms of the initial phase, previous studies have employed two distinct methodologies. First, the YTD18 ~\cite{makino2019recurrent} dataset of 30k hours of utterances (i.e. not publicly accessible), there is no explicit mention of their approach to video selection. Another approach utilized, as seen in the creation of the publicly accessible and extensively utilized LRS3 \cite{afouras2018lrs3}, involves constraining content sources to high-quality videos, specifically focusing on TED and TEDx talks. These talks are chosen due to their reliable transcripts and visuals that closely align with the targeted objectives, thereby maximizing the efficiency of a processing pipeline aimed at constructing a VSR dataset. However, this approach significantly reduces the pool of available content from YouTube and may restrict the dataset's capacity to represent real-world performances, as TED talks typically share similar visual contexts, featuring underrepresented individuals and employing a formal language vocabulary. We opt for an approach that combines the strengths of both strategies by focusing on keyword-based searches within a curated set of high-quality video sources. This hybrid approach allows us to expand the dataset's diversity while still ensuring the content's relevance and quality. By targeting videos that are keyword-tagged with specific themes and subjects, we enhance our ability to include a broader array of visual contexts and linguistic styles. This method not only diversifies the visual and linguistic input but also broadens the demographic representation within the dataset. Additionally, this strategy mitigates the limitations imposed by exclusively using TED and TEDx talks, thereby providing a more comprehensive foundation for developing robust and effective audio visual speech recognition models.

In this collection pipeline, and to avoid the complexity of large-scale processing, we aim to  ensure that the initially targeted videos are likely to contain VSR content. For this, we train a simple binary classifier that streams the first 100 frames and assigns a score whether the input should be further considered for processing. Thus, we reduce the search to approximately only 20 \% of the initial pool. The training set used to create the classifier was built by doing a first pass and then recursively tracking the videos that did not output any clips, followed by labeling the data accordingly.

\subsection{Data Gathering}

We leverage the capabilities of the YouTube search API, which provides various filtering options: (\textit{i}) Search using keywords, (\textit{ii}) target most relevant videos in regard of a specified language (Here French, Spanish, Arabic and Chinese). Indeed, we initiate the search using a set of $200$ keywords, such as "interview" or "discussion," to acquire the most pertinent content, extending the content-oriented filtering methodology explored in \cite{afouras2018lrs3}. Additionally, we ensure that these videos do not duplicate content from existing multi-lingual datasets (i.e. VoxCeleb2~\cite{nagrani2020voxceleb} and AV-Speech~\cite{ephrat2018looking}) by cross-referencing YouTube IDs. The subsequent videos are then processed as detailed next.

\subsection{Data Processing}

Each video is divided into multiple shots using a scene change detection~\cite{7040826}, which relies on alterations in three-dimensional histograms. Faces within each frame of the video are identified using YOLOv5n0.5-Face~\cite{qi2023yolo5face}, chosen for its high accuracy-to-compute cost ratio. Then, the detected faces are matched and tracked across frames to generate multiple face tracks. These tracks are then filtered using SyncNet ~\cite{chung2017out}, leveraging active speaker detection to isolate face track segments featuring speakers corresponding to the audio content. Finally, we utilize the ASR model Whisper ~\cite{radford2022robust} to detect the language, also obtaining automated-transcripts. Utilizing word timestamps from Whisper outputs, tracks are segmented into clips ranging from $2.0$ to $16$ seconds in duration.

\subsection{ViSpeR Statistics}

\textbf{Training:} As shown in Table \ref{data_stats}, our proposed dataset surpasses others in scale and coverage. ViSpeR exhibits substantial increases in both the number of clips and the total duration across all languages, making it a comprehensive resource for non-English VSR research.

\noindent\textbf{Test:} To ensure fair and robust evaluations, we take additional measures. Firstly, we obtain a second transcription using the Seamless-M4T model \cite{pratap2023scaling} for a pool of considered samples to build the test sets. Then, we retain only the clips that match the transcripts generated by Whisper, thus ensuring the creation of high-quality and reliable test sets. Additionally, we curate a subset from both the TedX and Wild splits. This ensures that our evaluation is both thorough and consistent with the LRS3 English TedX benchmark. For English, we use our previously introduced benchmark WildVSR \cite{djilali2024vsr} that is challenging and gives an accurate estimate of how existing English VSR models perform in the wild.

\begin{table}[t] 

    \centering
    \caption{\textbf{Comparison of VSR datasets}. Our proposed ViSpeR dataset is larger in size compared to other datasets that cover non-English languages for the VSR task. For our dataset, the numbers in parenthesis denote the number of clips. We also give the clip coverage under TedX and Wild subsets of our ViSpeR dataset.\vspace{-0.3cm}}
    \label{tab:stats_compare}

    \setlength{\tabcolsep}{4pt}
    \adjustbox{width=1\columnwidth}{
    \begin{tabular}{l| c |c| c |c}  
        \toprule[0.1em]
        
            \textbf{Dataset}  & \textbf{French} (\textit{fr}) & \textbf{Spanish} (\textit{es}) & \textbf{Arabic} (\textit{ar}) & \textbf{Chinese} (\textit{zh}) \\
            \midrule
            MuAVIC      & 176  & 178 & 16 & -- \\
            VoxCeleb2      & 124  & 42 & -- & -- \\
            AVSpeech      & 122  & 270 & -- & -- \\
            \midrule
          \textbf{ViSpeR} (TedX)  & 192 (160k) & 207 (151k)  & 49 (48k) & 129 (143k)   \\
          \textbf{ViSpeR} (Wild)  & 680 (481k) & 587 (383k)  & 1152 (1.01M) & 658 (593k)   \\
          \textbf{ViSpeR} (full) & 872 (641k) & 794 (534k)  & 1200 (1.06M) & 787 (736k)   \\
            
        \bottomrule[0.1em]
    \end{tabular}
    \vspace{-0.7cm}
    }
    \label{data_stats}
\end{table}

\begin{table}[t]
\caption{\textbf{Test set size per language.} For both TedX and Wild splits, the duration is given in hours. The numbers in parenthesis denote the number of clips.\vspace{-0.3cm}}
\centering
\begin{tabular}{l|cc}

\toprule
 & \textbf{TedX} & \textbf{WildVSR}\\
% \midrule 
\specialrule{0.5pt}{0.5pt}{1pt}

{French} (\textit{fr}) & 0.31 (221) & 2.01 (1442)  \\
{Spanish} (\textit{es}) & 0.65 (429) & 1.21 (828)  \\
{Arabic} (\textit{ar}) & 0.26 (208) & 1.19 (745)  \\
{Chinese} (\textit{zh}) & 0.37 (387) & 3.30 (2989)  \\
 \bottomrule
\end{tabular}

\end{table}

\section{Experiments}
\subsection{Experimental Setup} The processed multilingual VSR video-text pairs are utilized to train a encoder-decoder model in a fully-supervised manner. The encoder-decoder model closely follows the structure of the state-of-the-art AutoAVSR~\cite{ma2023auto}. The models are trained under a multi-lingual setting. While the encoder size is 12 layers, the decoder size is 6 layers. The hidden size, MLP and number of heads are set to 768, 3072 and 12, respectively. 
% The audio encoder for the AVSR model follows the same architecture as the encoder of the VSR model. 
The unigram tokenizers are learned for all languages combined and have a vocabulary size of 21k.
% , while the multilingual tokenizer is learned jointly with a vocabulary size of 21000. 
The models are trained for 150 epochs on 64 Nvidia A100 GPUs (40GB) using AdamW optimizer with max LR of 1e-3 and a weight decay of 0.1. A cosine scheduler with a warm-up of 5 epochs is used for training. The maximum batch size per GPU is set to 1800 video frames.

\begin{table}[t]
\caption{\textbf{Performance comparison on different languages.} Here, our multilingual ViSpeR models (VSR and AVSR) are evaluated on the TedX and Wild test splits combined. For \textit{en}, we combine the LRS3~\cite{afouras2018lrs3} and WildVSR~\cite{djilali2024vsr} test sets. Both VSR and AVSR models are trained on the full set (TedX+Wild).  Performance (lower is better) is reported in terms of WER for \textit{en}, \textit{fr}, \textit{es} and \textit{ar}, while CER is used for \textit{zh}. \label{exp:mono_results}\vspace{-0.3cm}}
\centering
\begin{tabular}{l|c|c}

\toprule
& VSR & AVSR \\
\midrule 

French (\textit{fr})
 & 29.8 & 5.7 \\

Spanish (\textit{es})
 & 39.4 & 4.4 \\

Arabic (\textit{ar})
 & 47.8 & 8.4 \\

Chinese (\textit{zh})
 & 51.3 & 15.4 \\

English (\textit{en})
 & 49.1 & 8.1 \\

 \bottomrule
\end{tabular}

\end{table}

\subsection{Results} 

Table~\ref{exp:mono_results} shows the performance of multilingual VSR and AVSR models for five languages (\textit{fr}, \textit{es}, \textit{ar}, \textit{zh},  and \textit{en}) on our proposed benchmarks. In general, we observe that the model performance on Latin languages is better (lower WER) compared to the non-latin languages (\textit{ar} and \textit{zh}) on both VSR and AVSR tasks. For Arabic, a likely explanation is the diversity of accents and the dynamic spellings for the same words. In addition, since we used Whisper to transcribe the segments, we expect a higher level of label-noise in non-Latin languages given the fact that Whisper performance on these languages isn't on par with the Latin ones. Furthermore, the models perform better on the wild test split, compared to the TedX test split for all languages except English. This can likely be attributed to the following reasons: (\textit{i}) the number of clips in train set belonging to the wild are far higher (4-10x) than those from TedX, and (\textit{ii}) while the wild set generally covers high-quality clips with faces near to the camera, the quality of the clips in TedX (non-English TedX isn't the official organisation and the setting isn't as professional) is lower due to farther camera angles, thereby resulting in sub-optimal decoding of the text from noisy video features. 

Moreover, these baseline model performances are still on the higher side ($\geq$30 WER) when compared to the state-of-the-art models on English ($<$20 WER) on LRS3 (\ie, less than 1h of duration). This can be attributed to (\textit{i}) smaller training sets in the proposed ViSpeR (800 to 1200 hours per language), compared to English (1700 to 3450 hours in training) and (\textit{ii}) harder test sets in our ViSpeR dataset ($\geq$1.5 hours per language) compared to English LRS3 test set (0.9 hours). Additionally, the recent work of \cite{djilali2024vsr} showed that VSR models drop in performance when tested on the newly introduced English benchmark (WildVSR), which is consistent with our findings. 

Furthermore, the results presented in Table~\ref{exp:mono_results} illustrates a significant performance disparity between Audio-Visual Speech Recognition (AVSR) and Visual Speech Recognition (VSR) across all evaluated languages (French, Spanish, Arabic, Chinese, and English). This difference can be largely attributed to the integration of audio cues in AVSR, which significantly enhances the model's ability to predict spoken words, even in challenging conditions such as noisy environments or videos with suboptimal visual clarity.

\section{Discussion}
\noindent\textbf{Ethical Considerations:} 
Given that the data collection for building the ViSpeR dataset utilizes publicly available videos from YouTube, biases inherent on the platform are likely to be present in the collected training and test sets. While steps have been taken to enhance the diversity of the content, the data is likely to be non-uniform and skewed towards certain content types during the filtering. Although the content used to derive the dataset is publicly available, we will provide a mechanism for content creators to opt out of the dataset. If anyone wishes to have their data removed, they can contact us and we will promptly exclude the associated clips and update the dataset.

\noindent\textbf{Future works and why ViSpeR dataset is important:}

\noindent\textit{Self-Supervised Learning methods for VSR:} A likely future direction includes training multilingual self-supervised models (similar to AV-HuBERT~\cite{shi2022learning}) on the proposed dataset to create a foundational model for VSR. This will include finding suitable clustering methods for creating the pseudo-labels to account for the multi-lingual aspect.

\noindent\textit{Multi-lingual supervised models:} When training a `single-model-for-multiple-languages', a few important questions that arise include: What is the optimal vocabulary size of the tokenizer? How to avoid tokens switching when the model mixes languages in the prediction? How to predict the spoken language if not know \textit{a priori}?

\noindent\textit{VSR translation:} Another interesting question involves leveraging the dataset to train models for translating visual speech from one language to another, such as from French to English. This capability holds great potential for facilitating seamless communication across linguistic barriers.

\noindent\textit{Other applications:} Beyond VSR, our dataset could also be used for lip syncing, speaker identification, etc.

\noindent\textbf{Conclusion} We proposed a large-scale multilingual dataset called ViSpeR for the task of Audio Visual Speech Recognition. The ViSpeR dataset contains nearly 3.2 million clips with more than 3600 hours duration in total, covering four languages: Chinese, Arabic, Spanish and, French. The clips were filtered from different settings like interviews, talks, \emph{etc.} to ensure sufficient diversity in the dataset. Furthermore, the test set contains two splits (TedX and WildVSR) per language to aid effective evaluation of the trained models. Moreover, we trained  multi-lingual baseline models in a fully-supervised manner on the ViSpeR dataset for VSR and AVSR. We observed a reasonable  performance on our proposed benchmarks, with clear gap between VSR and AVSR. 

{
    \small
    \bibliographystyle{ieeenat_fullname}
    \bibliography{main}
}


\end{document}